# Integrating Generative Artificial Intelligence with Learning Style Models: Opportunities, Challenges, and Future Directions

Malithi Nawarathne*, Prasanna Haddela, Nuwan Kodagoda

*Faculty of Computing, Sri Lanka Institute of Information Technology, Sri Lanka*

**Abstract -** Generative Artificial Intelligence (Generative AI) has emerged as a powerful tool for transforming learning. This review examines how Generative AI can be integrated with established learning style models (VARK, Kolb, Felder-Silverman, and Honey & Mumford) to support student engagement in large online classes. Following PRISMA 2020 guidelines, we searched Scopus, IEEE Xplore, and Google Scholar (2015–2026) and included 41 studies that met the eligibility criteria. The review maps Generative AI capabilities to learning style preferences, synthesizes existing implementations, highlights benefits and challenges, and proposes pathways for future research and practice. Despite ongoing debates regarding the empirical validity of learning styles, Generative AI enables flexible, multimodal, and real-time adaptation. This study offers practical insights for educators and developers seeking to build more effective, learner centered learning systems.

## 1. Introduction

The recent emergence of Generative Artificial Intelligence (Generative AI) has unlocked powerful capabilities that are transforming education (Baidoo-Anu & Ansah, 2023; Murtaza et al., 2024; Pesovski et al., 2024). Generative AI now supports intelligent course design, real-time assessment, tutoring, interactive content generation, and adaptive learning experiences (Bender, 2024; Biswas, 2023; Chan & Lee, 2023; Pesovski et al., 2024; Sharma & Yadav, 2022; Yu & Guo, 2023). In particular, Generative AI excels at delivering immediate, scalable, and personalized learning by dynamically generating tailored content, feedback, and learning pathways based on individual learner data (Abdullah et al., 2024; Chiu, 2023; Hashmi et al., 2024; Oseremi Onesi-Ozigagun et al., 2024; Randver, 2025).

Traditional teaching methods, however, continue to deliver instruction to all students at the same pace and in the same format. This one-size-fits-all approach struggles to accommodate individual differences in learning styles, preferences, prior knowledge, and needs. Personalized learning addresses this limitation by tailoring educational experiences to each learner's unique profile, thereby enhancing motivation, engagement, efficiency, and overall satisfaction (Ambele et al., 2022; Asthana et al., 2023; Grant & Basye, 2014; Keefe, 2007; Shemshack & Spector, 2020). The integration of Generative AI further strengthens personalization by enabling immediate, context-aware feedback and adaptive content delivery (Chiu, 2023; Oseremi Onesi-Ozigagun et al., 2024).

Despite growing interest, researchers have highlighted that substantial work remains to fully realize personalized learning at scale (Asthana et al., 2023; Chiu, 2023). Established learning style models such as VARK, Kolb, Felder-Silverman, and Honey & Mumford offer practical frameworks for understanding learner differences and guiding personalization efforts.

Nevertheless, despite rapid Generative AI adoption in education, no comprehensive synthesis currently exists that systematically links Generative AI capabilities with these established learning style frameworks for large-scale online personalization. This review addresses that critical gap by examining how Generative AI can be effectively integrated with learning style models to support scalable, learner-centered education. We now turn to why personalized delivery via learning styles is both necessary and underexplored, where this review was guided by the following research questions:

1. How can Generative AI capabilities be mapped to established learning style models (VARK, Kolb, Felder-Silverman, and Honey & Mumford)?
2. What implementations of Generative AI integrated with learning style models currently exist?
3. What are the benefits, challenges, and limitations of this integration?
4. What future directions should be pursued for research and practice?

This study followed a systematic review approach to identify and analyze research at the intersection of Generative AI and learning styles for application in large online classes. We searched major databases, including Scopus, IEEE Xplore, and Google Scholar, using combinations of keywords of “Generative AI”, “learning styles”, “virtual assistants”, “teaching assistants”, and “personalized learning”. Hence, studies were included if they met the following requirements,

1. Articles that were peer-reviewed journal articles, or conference papers published in English.

2. Articles focused on the application or integration of Generative with at least one established learning style model (VARK, Kolb, Felder-Silverman, Honey & Mumford).

3. Articles that addressed personalization, adaptive learning, teaching assistants, and virtual assistants in educational contexts.

4. Articles that were published between 1st of January 2015, and 31st of March 2026.

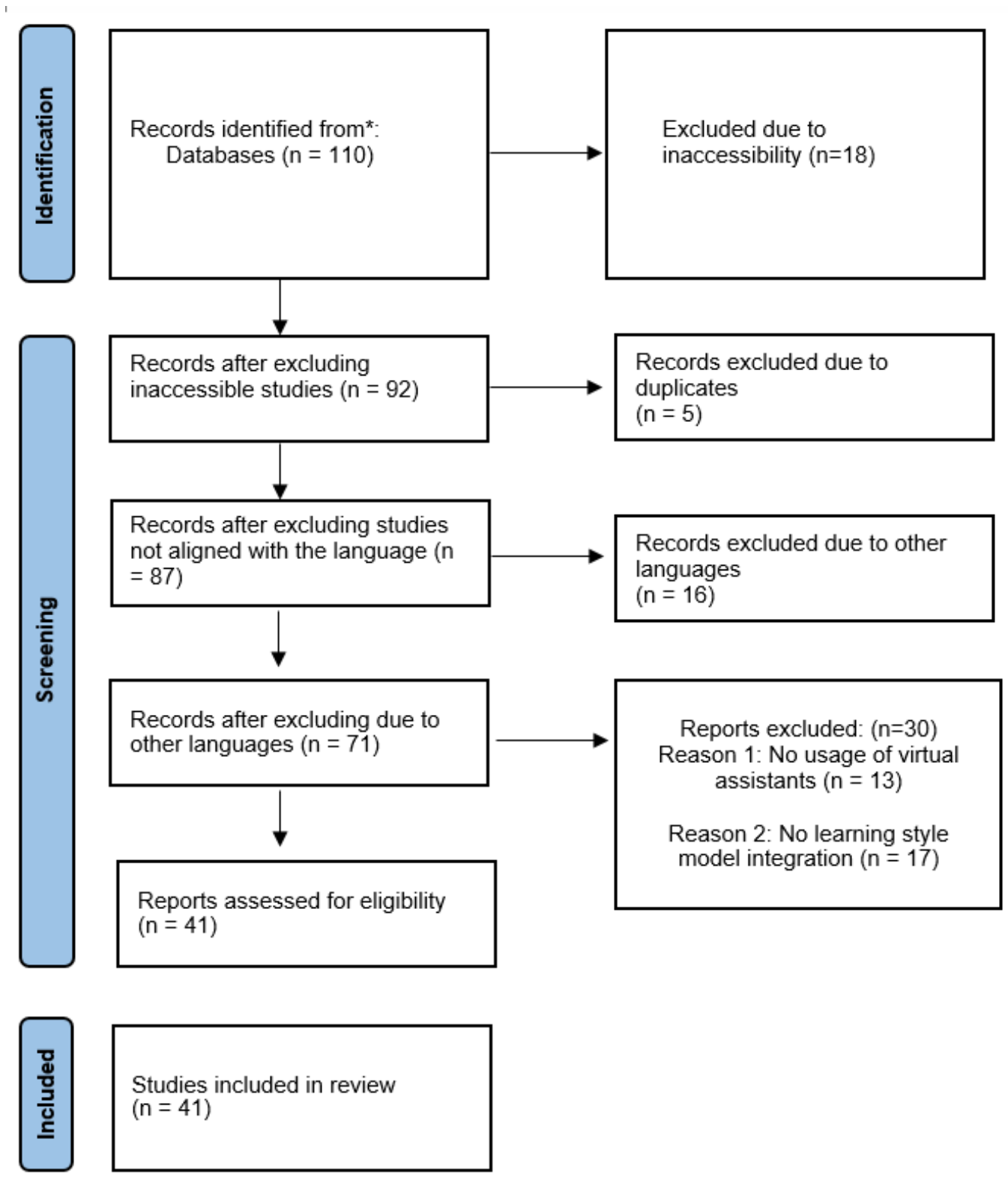

Figure 1: Prisma diagram related to the study

## 2. Learning Styles

Learning styles are preferred methods for processing information (Alzain et al., 2018; Khanal et al., 2014). Aligning teaching with styles improves outcomes, engagement, and efficiency (Abdullah et al., 2024; Budiyanto et al., 2017; Mahmoud Hawa & Tılfarlıoğlu, 2019). Further, aligning teaching with preferred styles moves beyond "one-size-fits-all" approaches, enhancing efficiency, problem-solving, and long-term achievement (Abdullah et al., 2024; Budiyanto et al., 2017). It addresses individual differences, boosts motivation, and is crucial as class sizes grow and technology evolves (Alhathli et al., 2017; Mahmoud Hawa & Tılfarlıoğlu, 2019).

However, although the concept of learning styles has faced significant criticism for lacking robust empirical evidence (Newton & Salvi, 2020; Pashler et al., 2008), it continues to inform practical research in adaptive and AI-driven education (Castro et al., 2025; Nunez et al., 2026; Randver, 2025; Ren et al., 2026, p. 2026; Troussas et al., 2025). Recent studies argue that, despite theoretical debates, learning styles frameworks enable dynamic personalization in technology-mediated environments, where Generative AI can generate multimodal content to accommodate preferences without rigid matching (Chandrasekera et al., 2024; Cohen, 2023; Fernando & Premadasa, 2024; Nunez et al., 2026; Oseremi Onesi-Ozigagun et al., 2024; Ren et al., 2026). Rather than framing Generative AI's objective as matching a predefined style, we emphasize its capacity for multimodal flexibility. By offering the same content in multiple formats at once, Generative AI allows students to select what works best for their particular context, avoiding the rigidity of forcing all learners into a single stylistic mode.

Therefore, this review was conducted with the aim of cautioning against over-reliance on styles as fixed traits while exploring Generative AI's potential to leverage them for flexible, engaging adaptation in large online classes. Hence, we have used Table 1, which compares several established learning style models in terms of their strengths and limitations, providing a foundation for evaluating the enhancements of Generative AI. While some emphasize sensory modalities (VARK), others focus on cognitive processing (FSLSM), experiential cycles (Kolb), or behavioral tendencies (Honey & Mumford).

**Table 1 – Learning style models comparison**

| **Model** | **Core Dimensions** | **Learning Focus** | **Strengths** | **Criticisms / Limitations** |
|---|---|---|---|---|
| VARK | Visual, Auditory, Read/Write, Kinesthetic | Sensory preferences | Improves understanding, desire, and engagement (Othman & Amiruddin, 2010); simple model (Rinekso, 2021); aids faster skill acquisition, positive attitude, and knowledge grasp (Sockanathan & Tajuddin, 2024). | Relies on self-reports (Oladele, 2024); may not capture the full complexity of preferences (Tomić et al., 2023). |
| FSLSM | Active/Reflective, Sensing/Intuitive, Visual/Verbal, Sequential/Global | Cognitive preferences | Provides complete descriptions, categorizing styles into detailed groups, unlike simpler (Hu et al., 2021). | Assumes preferences directly lead to behaviors (Felder & Spurlin, 2005). |
| Kolb | Concrete Experience, Reflective Observation, Abstract conceptualization, Active Experimentation | Learning as an experience cycle | Cycle enables feedback for new actions/evaluations (Almeida & Mendes, 2010); improves movement across cycle for better effectiveness (Kolb & others, 2007). | Oversimplifies process (Villanueva, 2020); lacks social/cultural context (McLeod, 2017). |
| Honey & Mumford | Activist, Reflector, Theorist, Pragmatist | Learning preferences derived from Kolb | Probes attitudes/behaviors for personal/organizational development (Trifirò & Laing, 2021); offers insights into | Lacks statistical pedagogical impact (Idowu, 2024) (Idowu, 2024).; low |
| | | | information engagement/ (Idowu, 2024). | validity/reliability (Duff & Duffy, 2002). |

As summarized in Table 1, each learning style model offers a different approach but also has limitations. Therefore, the comparative analysis shows that, with respect to each model, different personalization options can be offered as AI-based tools and technologies advance. A promising aid for operationalizing this integration is the use of virtual assistants and intelligent tutoring systems. The following section examines how virtual assistants have evolved from traditional AI-based tools to Generative AI powered systems and how they can effectively incorporate learning style models to deliver scalable personalized learning experiences in large online environments.

## 3. Teaching Virtual Assistants from Past to Present

Virtual assistants have functioned in organizing the study process, managing time, increasing motivation, and engagement in the study process (Gubareva & Lopes, 2020), (Pereira et al., 2022). These tools with the usage of AI, can provide personal assistance to students in the form of feedback, suggestions, scaffolding, and information retrieval. Before the introduction of Generative AI, many virtual teaching assistants were created with the use of AI based tools and technologies (Benedetto et al., 2019; Escotet, 2023; Essel et al., 2022; Liu et al., 2024; Mekni et al., 2020). However, the learning process can greatly benefit from Generative AI's integration of real-time feedback, adaptive material distribution, and powerful analytics.

Hence, studies on personalization and adaptive learning highlight Generative AI's role in tailored content, feedback, and engagement (Chan & Hu, 2023; Chen et al., 2023; Liu et al., 2024; Michel-Villarreal et al., 2023; Mittal et al., 2024; Wang et al., 2024). While these studies offer important findings and information, they also have several limitations in implementing personalized learning solutions via learning styles that can adapt according to changing student preferences. In order to provide personalized, flexible solutions, this combination pinpoints the necessity for research that combines Generative AI with pedagogical approaches. These systems can adapt to the demands of students in real time feedback. Summarizing these studies highlights both the progress made and the gaps that remain, particularly in how Generative AI can extend or improve upon earlier approaches. Tables 2,3,4,5 depict existing learning systems and virtual assistants that have been built using AI and ML technologies.

**Table 2 – Existing work done using FSLSM with AI-ML and Generative AI technologies**

| **Research** | **Contribution** | **Technologies/ Instruments used** |
|---|---|---|
| (Zagulova et al., 2019) | Linked FSLSM styles to performance, recommending personalized e-learning. | Index of learning style questionnaire |
| (Tzenios, 2020) | Used unsupervised machine learning techniques for health education grouping. | K-Means clustering |
| (Sihombing et al., 2020) | Applied FSLSM to develop personalized e-learning content. | ADDIE modelling |
| (Sensuse et al., 2020) | Conducted a literature review to show findings and recommendations to build an adaptive e-Learning system based on FSLSM. | Systematic literature review |
| (Supangat & Mohd Zainuri, 2020) | Tailored content using FSLSM. | Index of learning style questionnaire |
| (Boussaha & Drissi, 2021) | Proposed collaborative coaching using the FSLSM model. | Index of learning style questionnaire |
| (Rishard et al., 2022) | Developed “Adaptivo” that provides a personalized learning experience to the learners based on their learning style and knowledge level. | Index of learning style questionnaire, K-Means clustering, Random Forest, SVM, Decision Tree, and Logistic Regression |
| (Saputra et al., 2023) | Used SVM and improved the prediction accuracy of student learning styles | Support Vector Machine (SVM) algorithm |
| (Zagulova et al., 2023) | Discussed the possibility of using FSLSM for distance personalized learning. | Hypothesis testing |
| (Andriani, 2023) | Correlated styles to online achievements through FSLSM. | Index of learning style questionnaire |
| (Hussain et al., 2024) | Combined deep learning with sentiment analysis by combining automated learning style recognition using FSLSM. | Fuzzy logic |
| (Baihaqi et al., 2024) | Formulated a novel predictive framework based on for identification of student learning styles. | Artificial Neural Network (ANN) |
| (Randver, 2025) | Used FSLSM to prove that personalized learning improves student engagement, problem solving skills, and fosters metacognitive awareness. | Conceptual / Empirical study |
| (Lhafra & Abdoun, 2025) | Used clustering with the FSLSM to provide effective adaptive e-learning. | K-means |
| (Fernando & Premadasa, 2024) | Combined FSLSM with Generative AI to prove that fine-tuning education experiences, can allow personalization. | Fuzzy inference system (FIS) |
| (Vaccaro Jr et al., 2024) | Explored how LLMs can adapt science texts considering the learning preferences of middle school students using FSLSM. | GPT-4 |

**Table 3 – Existing work done using VARK model with AI-ML and Generative AI technologies**

| Research | Contribution | Technologies/ Instruments used |
|---|---|---|
| (Peter et al., 2009) | Demonstrated the use of the VARK model in adaptive e-learning to provide personalized materials tailored to learners' styles. | Semantic web technologies that include XML, RDF |
| (Sayed et al., 2020) | Used reinforcement learning with the VARK in predicting learning styles. | Multitask deep Q-learning, Bloom's taxonomy |
| (Rao & Arunachalam, 2021) | Used ensemble methods with the VARK model to predict learning styles. | J48, SVM, Naive Bayes and Random forest |
| (Pardamean et al., 2022) | Developed and measured the impact of an AI based learning style prediction model in an online learning portal for primary school students. | Matrix factorization |
| (Altamimi et al., 2022) | Used the VARK model to make learning styles predictions comparing regression and classification techniques. | Regression models and classification models |
| (Vanegas et al., 2024) | Designed a virtual course that adapts to different learning styles according to the | Surveys and interviews |
| | VARK model and evaluates students' academic performance and satisfaction. | |
| (Abomelha & Newbury, 2024) | Used the VARK model to build a recommendation system to identify learning styles that provided a personalized learning experience to students. | Fuzzy logic |
| (Nguyen et al., 2024) | Experimented on personalizing course content and learning activity model tailored for online courses based on students' learning styles using VARK. | Support vector machine |
| (Chandrasekera et al., 2024) | Investigated how Generative AI (Generative AI) can accommodate the various learning preferences defined in the VARK model (Visual, Auditory, Reading/Writing, and Kinesthetic) in design education | Technology acceptance (TAM) |
| (Oseremi Onesi-Ozigagun et al., 2024) | Experimented on how Generative AI can transform education by integrating the VARK model. | Conceptual / Empirical study |
| (Gunawan et al., 2024) | Explored on identifying factors in ChatGPT that could improve the learning process using the VARK learning style preferences | Systematic literature review |
| (Hashmi et al., 2024) | Investigated the impacts of Generative AI on confidence and frustration levels in programming among business students. | ChatGPT-4, K-Means |

**Table 4 – Existing work done using Kolb's model with AI-ML and Generative AI technologies**

| **Research** | **Contribution** | **Technologies/ Instruments used** |
|---|---|---|
| (Rasheed & Wahid, 2021) | Kolb's model was used to detect learning styles using classification techniques. | Decision trees, Support Vector Machines, K-Nearest Neighbour, Naïve Bayes, Linear Discriminant Analysis, Random Forest, and Logistic Regression |
| (Sanjabi & Montazer, 2020) | Applied personalization to e-learning environments using Kolb's model, finding that tailored learning improved academic performance and satisfaction. | Kolb's questionnaire |
| (Waladi et al., 2023) | Identify students having similar traits using Kolb model and proposed an intelligent adaptive learning system. | K-means, artificial neural network, decision tree |
| (Chaimae et al., 2024) | Employed the Kolb's model to study learning styles considering both balanced and imbalanced datasets. | KNN |
| (Sun & Deng, 2024) | Researched the use of ChatGPT to enhance experiential learning among college students using Kolb's Experiential Learning Model. | ChatGPT |
| (Özdemir et al., 2024) | Assessed the effectiveness of Generative AI tools in determining learning styles of students using Kolb's model | Kolb Learning Style Inventor, ChatGPT, Gemini, and Copilot |
| (Salinas-Navarro et al., 2024a) | Tested on the integration of ChatGPT with experiential learning introduced by | ChatGPT 3.5 |
| | Kolb to conduct feasible evaluations in applications of higher education. | |
| (Salinas-Navarro et al., 2024b) | Explored the use of Generative AI in developing learning activities using experiential learning for practical assessments. | ChatGPT 3.5 |
| (Luo et al., 2024) | Incorporated Kolb's experiential learning model along with LLMs to experiment on its usage to promote deep and high order cognitive skills. | Conceptual / Empirical study |

**Table 5 – Existing work done using the Honey and Mumford model with AI-ML and Generative AI technologies**

| **Research** | **Contribution** | **Technologies/ Instruments used** |
|---|---|---|
| (Sangvigit et al., 2012) | Examined the correlation between Honey and Mumford learning styles and students' preferences for online learning media using data stored in an LMS. | Learning Style Questionnaire of Honey & Mumford |
| (Morozevich et al., 2022) | Used Honey and Mumford's method of determining activity styles to develop a model for personalized learning paths using machine learning methods. | Matrix factorization and neural networks |
| (Lang, 2023) | Conducted a study to prove that student behavior on a LMS can be predicted using the Honey & Mumford Learning Styles Model questionnaire. | Learning Style Questionnaire of Honey & Mumford<br>Descriptive statistical analysis and hierarchical clustering |
| (Weijers et al., 2024) | Investigated the integration of Honey and Mumford's learning styles with LLMs to improve teaching effectiveness | GPT-4 |

Considering the existing work that has been done, it is evident that Felder-Silverman and VARK models have been used widely to propose solutions with the aid of AI related technologies. Kolb's model has been used mainly with the experiential learning concept in research works. Honey and Mumford model, however, has been used at a rare level.

Further, the evidence in tables 2, 3, 4, and 5 illustrates that while AI and ML techniques such as clustering, predictive modeling, and adaptive systems have successfully incorporated learning style models, FSLSM, VARK, Kolb, and Honey and Mumford, these implementations are often limited to static matching of learners to predefined content. Generative AI, by contrast, allows real-time adaptation, multimodal output, and conversational interaction, which can overcome the rigidity of earlier systems. Thus, the integration of Generative AI with these models represents a natural evolution beyond past personalization strategies.

## 4. Generative AI and Learning Style Models

### 4.1. Mapping Generative AI Tools to Learning Style Models

As observed in existing work, more studies could be conducted in mapping the learning styles with Generative AI tools to provide solutions. Building on the theoretical foundations of learning styles and the capabilities of Generative AI, the following mappings illustrate how specific content can be aligned with learner preferences. Unlike descriptive models, these mappings provide actionable insights for educators seeking to operationalize personalization in practice. Tables 6-9 outline that Generative AI can be used to map the learning styles.

The VARK model, which emphasizes sensory preferences (visual, auditory, read/write, and kinesthetic), has seen increasing attempts at integration with GenAI. Studies show that GenAI tools can effectively accommodate these modalities through multimodal generation (Chandrasekera et al., 2024; Gunawan et al., 2024; Hashmi et al., 2024; Oseremi Onesi-Ozigagun et al., 2024).

**Table 6 – Mapping VARK model with Generative AI tools based on evidence presented by** (Chandrasekera et al., 2024; Gunawan et al., 2024; Hashmi et al., 2024; Oseremi Onesi-Ozigagun et al., 2024)

| **Learning style mode** | **Preferred learning tools and materials** | **Mapping with Generative AI Tools** | **Learning benefit** |
|---|---|---|---|
| Visual | Graphs, charts, color coding, visualizations | Stable Diffusion for info graphics, flowcharts, and visual concept maps | These tools help clarify abstract topics by creating visual anchors that support spatial reasoning |
| Auditory | Discussion groups | LLMs with text-to-speech for narrated explanations or simulated conversations | This supports students who process best through listening and verbal engagement |
| Reading | Focusing techniques in writing, demonstrations, role-playing, open-ended questions | LLMs for detailed summaries, notes, quizzes, textual explanations | Promotes students who prefer reading to engage with materials that will improve their comprehension skills |
| Kinesthetic | Interaction with objects, internships, field trips, simulations, games | LLMs for scenarios/simulations, interactive step-by-step guides | Generate interactive step-by-step guides for physical tasks or projects |
| Multimodal | Lectures, notes, summaries, exams, presentations | LLMs for text; Stable Diffusion for visuals; speech-to-text for audio; virtual labs/scenarios | Supports students to learn via multimodal modes |

GenAI's strength lies in its ability to provide real-time adaptation across all eight dimensions of the Felder-Silverman model as researched by (Fernando & Premadasa, 2024; Vaccaro Jr et al., 2024).

**Table 7 – Mapping model with FSLSM Generative AI tools based on evidence presented by** (Fernando & Premadasa, 2024; Vaccaro Jr et al., 2024)

| **Learning style mode** | **Preferred learning tools and materials** | **Mapping with Generative AI Tools** | **Learning benefit** |
|---|---|---|---|
| Active | Exercises, discussion links | LLMs for real-time simulations, group problem-solving, interactive quizzes | Encourages application through active engagement |
| Reflective | References, PDFs, PPTs | LLMs for personalized reflection prompts, explanations, feedback | Promotes critical thinking and quiet information processing |
| Sensory | Practical materials | LLMs for real-life case studies, virtual labs simulating industry tasks. | Support experiential learning |
| Intuitive | Advanced topics, references, assignments, topic list, video, forum, PPTs, PDF | LLMs for underlying theories, advanced prompts, conceptual discussions/maps | Encourages exploration of studies |
| Visual | Graphics and visualizations animations, video | Stable Diffusion for illustrations, slide-based content | Clarifies abstracts via visual anchors for spatial reasoning |
| Verbal | Textual materials, audio Solved exercises after the content | LLMs with speech synthesis for narrated lectures, example explanations, dialogues | Promotes conversational learning |
| Global | Chapter summary and navigation, case analysis activities | LLMs for thematic guides, narrative summaries, concept networks | Provides interconnected content for holistic views |
| Sequential | Solved exercises and videos after the content | Generative AI for stepwise paths, progressive assessments, and sequenced delivery | Allows methodical thinking |

Recent studies show promising applications, particularly in using LLMs to support the learning cycle of Kolb's model (Luo et al., 2024; Özdemir et al., 2024; Salinas-Navarro et al., 2024a, 2024b; Sun & Deng, 2024). GenAI tools can generate personalized scenarios for concrete experience, reflection prompts for reflective observation, theoretical explanations for abstract conceptualization, and interactive simulations for active experimentation. This integration helps overcome the limitations of traditional Kolb-based systems by enabling dynamic, real-time movement through the experiential cycle.

**Table 8 – Mapping Kolb's model with Generative AI tools based on evidence presented by** (Luo et al., 2024; Özdemir et al., 2024; Salinas-Navarro et al., 2024a, 2024b; Sun & Deng, 2024)

| **Learning style mode** | **Preferred learning tools and materials** | **Mapping with Generative AI Tools** | **Learning benefit** |
|---|---|---|---|
| Diverging | Brainstorming, internet based interactive exercises online discussing | LLMs for brainstorming prompts, alternative perspectives, simulated peer discussions | Encourage creativity through multiple options and problem reframing |
| Assimilating | Reading lectures, exploring analytical models, step-by-step tutorials, online reading | LLMs for summarizing articles, structured tutorials, analytical comparison charts. | Helps learners build structured understanding from abstract material |
| Converging | Interactive activities, trial/error learning, theoretical tutorials, online reading, experiments | Generative AI for problem-solving scenarios, real-life decision-making/troubleshooting simulations | Allows students to engage with work that are in an interactive manner |
| Accommodating | Simulations, observations, examples, laboratories, online discussions, team tasks | VR/AR tools, Generative AI-generated instructions, group projects | Support trial and error learning and action driven discovery |

Although less frequently studied compared to the other models, Honey and Mumford's learning styles (Activist, Reflector, Theorist, Pragmatist) show strong potential for GenAI integration. Weijers et al. (2024) explored the use of LLMs with this model to improve teaching effectiveness, demonstrating how GenAI can support activists through gamified and interactive activities, reflectors through journaling prompts, theorists through conceptual explanations, and pragmatists through applied case studies and practical scenarios.

**Table 9 – Mapping Honey and Mumford model with Generative AI tools based on evidence presented by** (Weijers et al., 2024)

| **Learning style mode** | **Preferred learning tools and materials** | **Mapping with Generative AI Tools** | **Learning benefit** |
|---|---|---|---|
| Activist | Brainstorming, problem-solving, group discussions, puzzles, competitions, lectures/reading | Fine-tuned LLMs for gamified modules, timed challenges, and live group activities | Promote high interaction and engagement |
| Reflector | Paired discussions, self-analysis/personality questionnaires, observing activities, feedback, coaching, interviews | LLMs for journaling prompts, scenario reviews, comparative responses | Enables formative feedback and simulated peer evaluation |
| Theorist | Group discussions, reading lists | LLMs for reading lists, model definitions, concept maps, explanatory essays, comparison tasks | Promotes group work and cooperation |
| Pragmatist | Case studies, problem-solving, discussions | LLMs for case studies, task-oriented scenarios, applied exercises | Tailors examples for immediate relevance |

As seen across the four learning style models, Generative AI tools exhibit the flexibility to support diverse learner preferences using adaptive content. This approach not only empowers educators to diversify their teaching strategies but also lays the groundwork for building intelligent virtual assistants or tutoring systems capable of tailoring the delivery of content in real time with personalization.

Hence, building on VARK, Felder-Silverman, Kolb, and Honey & Mumford models, this study proposes on aligning each style with suitable Generative AI tools. Such adaptive delivery can not only increase engagement and satisfaction but also foster deeper learning, particularly in online educational settings. Importantly, this personalized alignment creates new opportunities to reduce cognitive overload and empower students to take ownership of their learning process.

### 4.2. Enhancing the learning experience in large classes through Generative AI personalization

However, while learning style-based personalization offers significant pedagogical value, it addresses only part of the broader personalization challenge. To realize the full potential of Generative AI, it is critical to extend the scope beyond predefined typologies and toward a more holistic, context-sensitive view of learner diversity. While learning styles serve as an important first layer in personalizing education, modern learners exhibit more complex, evolving needs that go beyond sensory and cognitive preferences. Personalization in education must also accommodate factors such as student engagement, motivation, student isolation, providing student feedback, student-teacher interaction, and diverse knowledge levels. Hence, beyond aligning content with individual learning styles, Generative AI offers unique affordances for enriching the broader learning experience. This section explores several pedagogical dimensions, ranging from emotional engagement to individualized feedback, through which Generative AI can deepen personalization in education.

#### *4.2.1. Student engagement and motivation*

Engagement and motivation are often compromised in online learning due to passive instruction and a lack of personalization. Learning styles, when supported by Generative AI, can reignite engagement by presenting content in formats students naturally respond to. Generative AI enhances this by dynamically adapting the format and interactivity level, using reinforcement learning, feedback signals from learners' inputs to keep them motivated and involved (Baidoo-Anu & Ansah, 2023; Chan & Lee, 2023).

#### *4.2.2. Emotional state and cognitive load*

Online learners often experience emotional fatigue, confusion, or anxiety, particularly when instruction does not align with their preferences. Generative AI can monitor sentiment and adapt instruction not only to how learners prefer to learn, but also to how they're feeling while learning. AI generated real world simulations and storytelling tasks can build emotional resonance. LLMs can adjust tone, simplify content complexity, and provide emotionally reassuring feedback based on observed learner frustration (Baidoo-Anu & Ansah, 2023; Boussaha & Drissi, 2021). Through natural language understanding, Generative AI can act as a responsive tutor, identifying confusion or stress signals and adapting content or pacing accordingly.

#### *4.2.3. Reducing student isolation and strengthening interaction*

Large online classrooms often lead to student anonymity and a sense of disconnection. Generative AI, integrated with learning style insights, can foster personalized interaction and reduce isolation. Generative AI tools can also simulate group work experiences, design peer-mirroring, or recommend tailored discussion prompts to foster community for learners who might otherwise remain disengaged (Hashmi et al., 2024). Generative AI tools tailored using learning style inputs, can serve as simulated conversational partners (Baidoo-Anu & Ansah, 2023; Hashmi et al., 2024; Pesovski et al., 2024). This human like interaction restores some of the relational elements missing from traditional online learning environments, thereby increasing perceived support and reducing social isolation.

#### *4.2.4. Personalized feedback providing*

Feedback is critical to learning, yet in large-scale virtual settings, it is often delayed or generic. Generative AI tools can deliver immediate, personalized feedback that aligns with a student's preferred mode of processing (Chan & Lee, 2023, 2023; Michel-Villarreal et al., 2023). It can tailor its explanations to match student preferences. Tools such as ChatGPT, Gemini, have already proved to assist students track their goals and generate personalized visual learning analytics dashboards tailored to their learning style (Randver, 2025). Such alignment ensures feedback is not only timely but also cognitively resonant and reinforces student growth more effectively.

#### *4.2.5. Supporting diverse knowledge levels*

In large classrooms, the spread of prior knowledge and competencies is wide. Generative AI, informed by learning style models, can dynamically adjust task difficulty and presentation. Through diagnostic quizzes or AI driven knowledge graphs, students can be categorized not only by level but also by preferred learning mode, enabling tailored learning paths that scale effectively across hundreds or thousands of students (Chan & Lee, 2023; Michel-Villarreal et al., 2023; Mittal et al., 2024; Pesovski et al., 2024).

#### *4.2.6. Enhancing student-teacher interaction*

While Generative AI is often used by students, it can also personalize support for teachers, using learning style data to help instructors monitor which students are disengaged or struggling and how best to intervene (Chan & Lee, 2023). Teachers could receive Generative AI generated recommendations on grouping students with complementary learning styles for collaborative projects. Analytics dashboards can highlight mismatches between student preferences and performance, prompting pedagogical adjustments. This allows educators to maintain a personalized presence at scale, closing the feedback loop and enhancing the human element of online teaching.

## 5. Conclusion

Considering the propositions discussed in this section, it is evident that introducing personalized learning with the integration of Generative AI tools is a plausible solution that can assist students to learn contentedly. The integration of learning styles with Generative AI has shown great potential in transforming the education field. Generative AI can overcome the limitations introduced in existing traditional and online learning techniques as discussed in this study. Most importantly, the Generative AI’s ability to adapt to present circumstances and provide fast responses can allow the students and educators carry out their tasks efficiently. While this study highlights the potential of Generative AI in supporting personalized learning through established learning style frameworks, it is important to acknowledge certain limitations and ethical considerations. Learning style theories have been a topic that has received attention due to critics arguing about oversimplification. Ethical considerations should also be taken into account in the usage of Generative AI related tools, as the research might involve working with sensitive data. Further research should therefore be conducted while addressing the ethical considerations that would result in virtual learning environments, balancing automation with pedagogical intents. In conclusion, the combination of learning styles with Generative AI represents a huge step forward in personalized education. The path ahead is not without its challenges, but the potential will reward empowered learners and educators to perform their duties efficiently. This review hence lays a solid roadmap that can be considered for the development of systems that can adopt Generative AI and learning styles.

## Conflicts of interest declaration

All authors declare that they have no conflicts of interest

## Declaration of AI Use

During the preparation of this work, the authors used Grok 4 to paraphrase and improve readability of sections. The authors reviewed, edited, and take full responsibility for the content.